\documentclass{article}

\usepackage{PRIMEarxiv}

\usepackage[utf8]{inputenc} % allow utf-8 input
\usepackage[T1]{fontenc}    % use 8-bit T1 fonts
\usepackage{hyperref}       % hyperlinks
\usepackage{url}            % simple URL typesetting
\usepackage{booktabs}       % professional-quality tables
\usepackage{amsfonts}       % blackboard math symbols
\usepackage{nicefrac}       % compact symbols for 1/2, etc.
\usepackage{microtype}      % microtypography
\usepackage{lipsum}
\usepackage{fancyhdr}       % header
\usepackage{graphicx}       % graphics
\usepackage{tikz}
\usepackage{subcaption}
\usepackage{amsmath}
\providecommand*\coloneqqf{=}

\title{FastPoseCNN: Real-Time Monocular Category-Level Pose and Size Estimation Framework
}

\author{
  Eduardo Davalos \\
  Vanderbilt University \\
  Nashville, TN \\
  \texttt{eduardo.davalos.anaya@vanderbilt.edu}\\
   \And
  Mehran Aminian\\
  St. Mary's University\\
  San Antonio, TX\\
  \texttt{maminian@stmarytx.edu} \\
}

\begin{document}
\maketitle

%%%%%%%%%%%%%%%%%%%%%%%%%%%%%%%%%%%%%%%%%%%%%%%%%%%%%%%%%%%%%%%%%%%%%%%%%%%%
\begin{abstract}
%\addcontentsline{toc}{chapter}{ACKNOWLEDGMENTS}

% Outline
% Source: https://writing.wisc.edu/handbook/assignments/writing-an-abstract-for-your-research-paper/

% 1. the context or background information for your research; the general topic under study; the specific topic of your research

The primary focus of this paper is the development of a framework for pose and size estimation of unseen objects given a single RGB image - all in real-time. % 
%
% 3. what’s already known about this question, what previous research has done or shown
%
In 2019, the first category-level pose and size estimation framework was proposed alongside two novel datasets called CAMERA and REAL. % 
%
% 4. the main reason(s), the exigency, the rationale, the goals for your research—Why is it important to address these questions? Are you, for example, examining a new topic? Why is that topic worth examining? Are you filling a gap in previous research? Applying new methods to take a fresh look at existing ideas or data? Resolving a dispute within the literature in your field? . . .
%
However, current methodologies are restricted from practical use because of its long inference time (2-4 fps). Their approach's inference had significant delays because they used the computationally expensive MaskedRCNN framework and Umeyama algorithm. %
%
% 5. your research and/or analytical methods
%
To optimize our method and yield real-time results, our framework uses the efficient ResNet-FPN framework alongside decoupling the translation, rotation, and size regression problem by using distinct decoders. Moreover, our methodology performs pose and size estimation in a global context - i.e., estimating the involved parameters of all captured objects in the image all at once. 
%
% 6. your main findings, results, or arguments
%
We perform extensive testing to fully compare the performance in terms of precision and speed to demonstrate the capability of our method.
%
% 7. the significance or implications of your findings or arguments.
%
% Our code will be made available in the following link: {\color{red}\href{https://github.com/edavalosanaya/FastPoseCNN}{https://github.com/edavalosanaya/FastPoseCNN}}.

\end{abstract}

% keywords can be removed

%%%%%%%%%%%%%%%%%%%%%%%%%%%%%%%%%%%%%%%%%%%%%%%%%%%%%%%%%%%%%%%%%%%%%%%%%%%%
\section{Introduction}
\label{ch:introduction}

% \subsection{Overview of Contents}

% The following thesis is divided into six major chapters. Chapter \ref{ch:introduction} - Introduction - begins our discussion about the field of pose and size estimation and demonstrates our proposed method. Chapter \ref{ch:background} - Background - provides a mathematical framework for 6D pose and size estimation and other supplementary equations used in our work. Chapter \ref{ch:literatureReview} - Literature Review - further specifies how this work fits within the literature and provides further context about 6D pose and size estimation. Chapter \ref{ch:methodology} - Methodology - dives into the details about our proposed model. Chapter \ref{ch:experimentsResults} - Experiments and Results - provides data describing the overall performance, limitations, and other key features of our proposed framework. The conclusion ends with elaborating on the future work and other possible improvements.

% \subsection{Pose and Size Estimation}

We study rigid-body 6D pose and size estimation to detect and recognize object's spatial information which includes translation, rotation, and scale. With all this information, we can pinpoint the object in 3D space and understand its relationship to the surrounding environment. This technology provides a foundation for various practical applications, including mixed reality, robotics, and object tracking. Historically, methods in 6D pose and size estimation were focused on feature matching through the use of algorithmic feature extractors such as SIFT (Scale-Invariant Feature Transform) \cite{Lowe2004_SIFT} and SURF (Speeded-Up Robust Features) \cite{Herbert2006_SURF}.

Pose and size estimation, similar to other computer vision (CV) tasks, has undergone a complete transformation with the emergence of deep learning and convolutional neural networks (CNN). In recent years, most proposed state-of-the-art methods have used end-to-end neural network models whose input is an image, and their output is the final pose and size of targeted objects.

% \subsection{FastPoseCNN}

\begin{figure}%[h]
    \centering
    \input{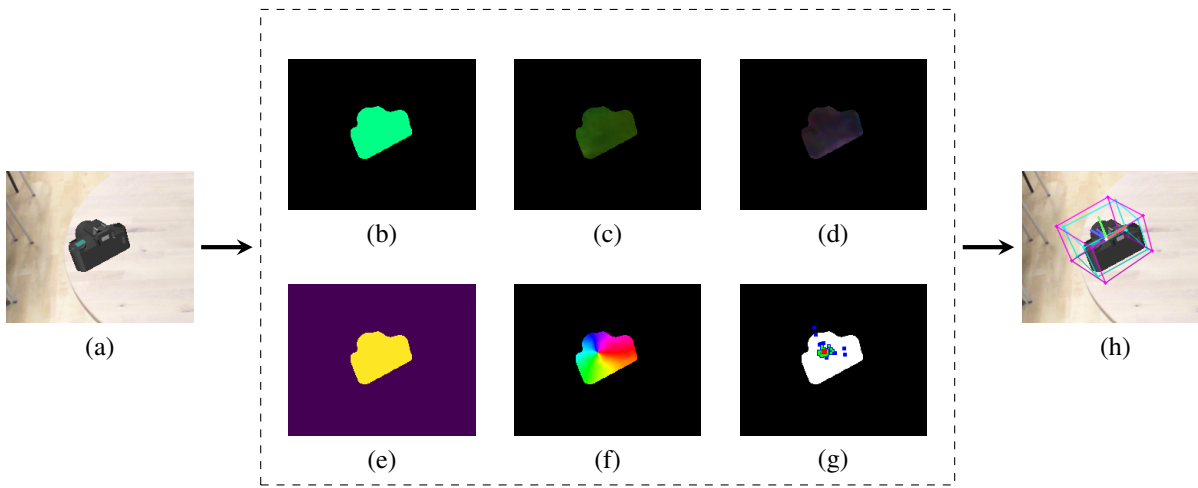}
    \caption{Model's Intermediate Data Representation. (a) Input RGB image, (b) segmentation, (c) quaternion, (d) size, (e) depth, (f) centroid vectors, (g) 3D centroid, (h) pose and size instance data.}
    \label{fig:data_representation}
\end{figure}
%
% Briefly describe our method
In this work, RGB images are used as input to make multiple dense pixel-wise predictions, including the centroid, quaternion, and size vector fields and depth regression. These dense outputs are converted into instance-wise attributes through an aggregation routine that allows the estimation of the final pose and size. The regress parameters are constructed from the collective information of an object's seen pixels through the use of pixel-wise data, resulting in robustness against occlusion and truncation. The code for this model which is open source can be found in the following GitHub repository: {\color{red}\href{https://github.com/edavalosanaya/FastPoseCNN}{https://github.com/edavalosanaya/FastPoseCNN}}.

% Provide our contribution's summary
The contribution of this work to the field can be summarized as follows:

\begin{itemize}
    \item Proposing a novel CNN model for category-level 6D pose and size estimation named FastPoseCNN. Our network generates dense pixel-wise predictions for each decoupled spatial component.
    \item Achieving real-time inference only requires RGB images as input, and it is robust to occlusion and truncation through its pixel-wise driven pose and size estimation.
    \item Introducing SymQuaternion-Loss, a new training loss function for quaternion regression that accounts for symmetric objects.
\end{itemize}

%%%%%%%%%%%%%%%%%%%%%%%%%%%%%%%%%%%%%%%%%%%%%%%%%%%%%%%%%%%%%%%%%%%%%%%%%%%%
\section{Literature Review}
\label{ch:literatureReview}

% Introduce the dividing features within the literature
The field of 6D pose and size estimation has many different approaches and methodologies for estimating rigid body spatial information. We begin our overview of the literature by an initial categorization from previous works - primarily the problem and its solution constraints. These constraints are illustrated in the following subsections: \ref{sec:data_type} specifies input data types to a model, \ref{sec:category_vs_instance} describes object variation, and \ref{sec:real_time_inference} elaborates the throughput requirement of solutions. Afterward, we will discuss the types of strategies and approaches used in the literature in Section \ref{sec:related_work} and how these methods compare to each other in terms of performance, efficiency, and usability. Finally, we conclude our literature review by addressing the gap mentioned in Section \ref{sec:literature_gap} using our method.

\subsection{Input Data Type Differences} \label{sec:data_type}

% \begin{figure}[h]
%     \centering
%     \input{Figures/literature_review/pose_estimation_method_categories}
%     \caption{Input Data Types}
%     \label{fig:pose_estimation_method_categories}
% \end{figure}

% Introduce the different types of input
The first distinction among many proposed solutions is the type of input image or information they used. Standard data inputs include RGB images, RGBD images, depth images, point clouds, and other input representations. Different data representations provide unique challenges, complexities, and constraints. 

% Talk about RGB
\paragraph*{RGB Image Input.}
Methods that used RGB images have gained high attention, mostly due to the commonality of RGB cameras in modern devices. Determining the 6D pose and size of an object from a single RGB image is inherently more challenging than using an RGBD image. It is because the missing depth information introduces perspective ambiguity. Even for humans, an object's unknown depth results in scaling ambiguities. Therefore, making pose and size estimation especially difficult.

% Talk about RGBD
\paragraph*{RGBD Image Input.}
Historically, RGBD methods have been more commonly used in the 6D pose estimation field \cite{Brachmann2014_3DObjectCoordinates}. Methods that rely on RGBD input images utilize the 3D features to capture the pose of objects more accurately. These methods have achieved excellent performance by using neural networks and other machine learning approaches \cite{Pereira2019_MaskedFusion, Park2019_LatentFusion, Tian2020_R6OPELDF}. One of the downsides of this mode of input data is the expensive requirement of an RGB+depth camera.

% Talk about D images
\paragraph*{Depth Image Input.}
Pure depth methods rely on the volumetric information provided in a depth image to estimate the pose \cite{Sahin2018_CLPRDI, Tan2017}. These methods use lighter machine learning models such as random forest models to make their inference extremely fast, e.g., 2 ms for \cite{Tan2017}. These methods benefit from smaller yet efficient models as they require fewer training samples. The major drawback of these methods is their relatively lower performance, compared to RGB and RGBD methods, in more challenging datasets. The decreased performance is more visible when tested on the difficult LINEMOD-OCCLUDED and LINDEMOD-TRUNCATION datasets.

% Talk about Point-Clouds
\paragraph*{Point-Clouds Input.}
With the development of PointNet \cite{Charles2016_PointNet}, Point clouds for CV methods have gained interest primarily because of their powerful integration of depth in a native 3D space. By lifting RGBD images into a point cloud representation, DL models benefit more from the present depth information by learning complex 3D features \cite{Xu2019_W-PoseNet, Wang2019_DenseFusion, Gao2020_PoseRegressionPC, Chen2020_G2L-Net}. Fusion methods combine the learned features from the point cloud and RGB image to better estimate an object's pose.

\subsection{Category-Level vs. Instance-Level} \label{sec:category_vs_instance}
% \begin{figure}[h]
%     \centering
%     % \hfill
%     % \centering
%     % \subfloat[NOCS CAMERA: A Category-Level dataset \label{category_level_dataset}]{%
%     \subfloat[\label{fig:category_level_dataset}]{
%         \input{Figures/literature_review/noc_camera_real_objects}
%     }
%     \newline
%     % \hfill
%     % \subfloat[LINEMOD: An instance-level dataset \label{fig:instance_level_dataset}]{%
%     \subfloat[\label{fig:instance_level_dataset}]{
%          \input{Figures/literature_review/linemod_objects}
%     }
%     % \hfill
%     \caption{(a) Category-Level vs. (b) Instance-Level Dataset}
%     \label{fig:category_level_vs_instance_level_datsets}
% \end{figure}

\begin{figure}[h]
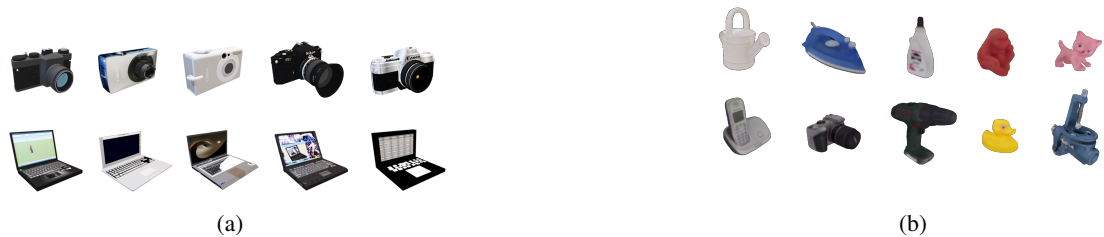

    \centering
    \begin{subfigure}[b]{0.45\textwidth}
        \centering
        \input{Figures/literature_review/noc_camera_real_objects}
        \caption{}
        \label{fig:category_level_dataset}
    \end{subfigure}
    \hfill
    \begin{subfigure}[b]{0.45\textwidth}
        \centering
        \input{Figures/literature_review/linemod_objects}
        \caption{}
        \label{fig:instance_level_dataset}
    \end{subfigure}
    \caption{(a) Category-Level vs. (b) Instance-Level Dataset}
    \label{fig:category_level_vs_instance_level_datsets}
\end{figure}

% Talk about instance-level datasets in how they were the first datasets used for pose estimation
The first modern datasets available for 6D pose estimation included LINEMOD \cite{Hinterstoisser2012_LINEMOD}, LINEMOD-OCCLUDED \cite{Brachmann2014_3DObjectCoordinates}, YCB-V \cite{Xiang2017_PoseCNN}, and T-LESS \cite{Hodan2017_T-LESS}. All of these datasets share the property of being instance-level datasets. Instance-level implies that the dataset does not include any variants for objects of the same type. There is only one instance of each type in an object class. This is a major flaw in SOTA pose research since it does not accurately reflect the nature of objects found in reality. The discrepancy between real-life data and SOTA research datasets has led to the need for a 6D pose and size category-level dataset.

% Then talk about NOCS and their new category-level dataset
The landmark publication by \cite{Wang2019_NOCS} provided the first pair of publicly available category-level datasets, named CAMERA and REAL. CAMERA is an extensive synthetic dataset containing realistic backgrounds with rendered 3D models in workplace-related environments. REAL is a smaller real dataset with the same object categories and similar backgrounds. These datasets included a variety of distinct instances for each object category, as shown in Fig. \ref{fig:category_level_vs_instance_level_datsets}. It added another level of complexity to pose estimation as now models had to account for these per-instance differences and how that might affect the effectiveness of methods.

\subsubsection*{Size Estimation Requirement}

\begin{figure}[h]
    \centering
    \input{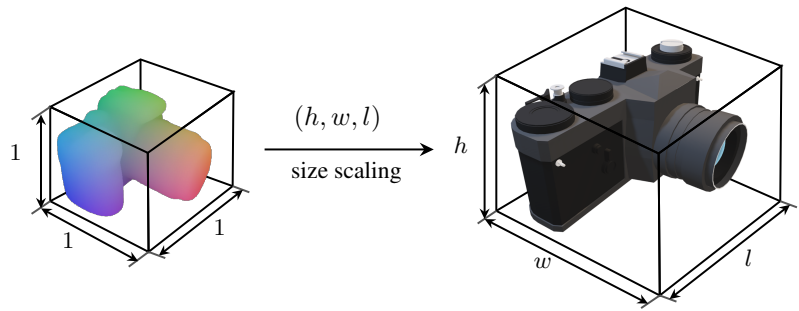}
    \caption{Size Regression}
    \label{fig:size_regression}
\end{figure}

% Talk about how category-level pose estimation adds the size requirement because of its nature
% - because it does not use exact 3D models
Additionally, \cite{Wang2019_NOCS} was the first paper to address category-level 6D pose and size estimation problem - specifically the addition of the size requirement. This is because instance-level pose estimation automatically provides the scale of an object since the exact 3D model of the object is known. However, the category-level problem adds the size requirement to fully estimate the complete bounding box of an object to account for per-instance size variations. 

\subsection{Real-Time Inference} \label{sec:real_time_inference}

% Talk about how real-time inference is wanted in applicable methods
% also elaborate on how NOCSNet is unsuccessful when it comes to this issue.
Another major concern present in pose and size estimation is the inference time of solutions. Models need to detect and classify objects and their corresponding pose efficiently to allow applications to be built on top of the model; therefore, this time efficiency requirement is a commonly sought feature in the CV. This was a problem for \cite{Wang2019_NOCS} as their proposed method was only able to run within 2-4 fps on an Intel Xeon Gold 5122 CPU @ 3.60GHz desktop with an NVIDIA TITAN Xp. The slow inference makes their MaskRCNN-NOCS model not practical for time-sensitive applications.

% Will only include the related work section to the actual journal paper
% but we include the previous literature review to provide more contextual information about 6D pose and size estimation.
\subsection{Related Work} \label{sec:related_work}

\begin{figure}[h]
    \centering
    \input{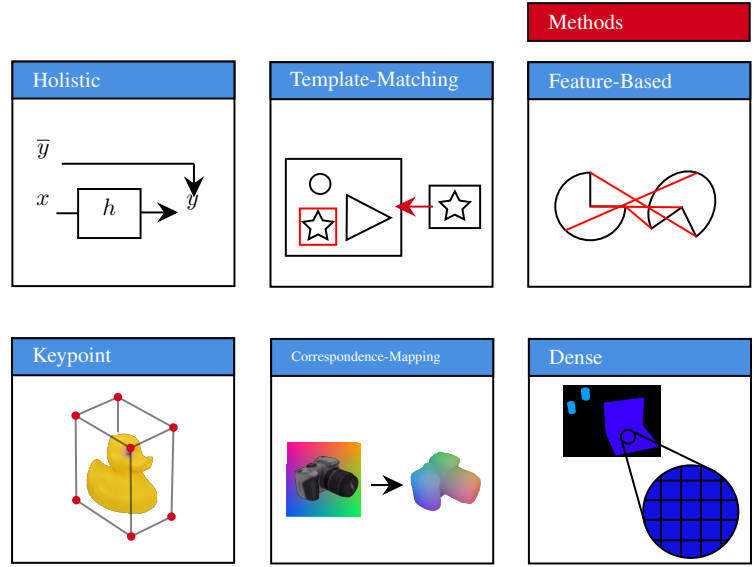} 
    \caption{Types of Methods - Simply Illustrated}
    \label{fig:pose_methods_illustrated}
\end{figure}

\paragraph*{Holistic Methods.}
% Explain how holistic methods are methods that perform direct regression/classification of 
% pose and size paramaters
Holistic methods take the approach of directly regressing the pose, size, or other object's attributes. To simplify the nonlinearity of the rotation space, \cite{Tulsiani2014_VK, Su2015_RenderCNN, Kehl2017_SSD-6D, Sundermeyer2019_Implicit} quantized the SO(3) space - making it into a more stable yet less accurate classification problem. It is common practice in the literature to take a mixed approach when estimating an object's 6D pose and size. \cite{Xiang2017_PoseCNN} uses a CNN feature extractor to estimate the decoupled translation and rotation. First, the extracted features were used to estimate the translation via dense keypoint regression to identify the centroid ($x,y$) and dense pixel-wise regression for the direct depth ($z$). Second, they used the extracted features to directly regress the rotation by approximating an object's quaternion $\bold{q}$.

\paragraph*{Template-matching Methods.}

% Talk about how templates used a sliding window fashion for finding objects and matching their pose.
Before the use of DL in pose estimation, template-matching was widely used to estimate the pose of rigid bodies \cite{Cao2016_TemplateMethod, Zhu2014_TemplateMethod, Chunhui2010_TemplateMethod, RiosCabrera2013_TemplateMethod, Huttenlocker1993_TemplateMethod, Hinterstoisser2011_TemplateMethod, Hinterstoisser2013_TemplateMethod, Hinsterstoisser2012_TemplateMethod}. Template-matching methods use a sliding-window algorithm that calculates a similarity score between an image and multiple perspective-based templates. The major advantage in template-matching is its great ability to estimate the pose of texture-less objects with great performance. However, its heavy reliance on the similarity score reduces its performance when exposed to occlusion, truncation, and lighting variations. 

\paragraph*{Feature-based Methods.}

% Talk about feature (both before and after DL introduction into the field)
Another method used in the traditional field of 6D pose estimation, hand-crafted, and feature engineering were used for feature extraction and matching \cite{Lowe1999_NonDLFeature, Collect2011_NonDLFeature, Rothganger2006_NonDLFeature}. However, feature extraction and matching require texture-rich objects to accurately detect and recognize these objects. With the help of CNN's in pose estimation, using trainable end-to-end neural network models has become commonplace as these approaches learn more effective methods for extracting features in more challenging scenarios \cite{Xiang2017_PoseCNN, Capellen2019_ConvPoseCNN, Hu2018_Segmentation, Zakharov2019_DPOD, Xu2019_W-PoseNet,Wang2019_NOCS, Wang2019_DenseFusion,Tremblay2018_DOPE}. After using features instead of templates, methods have become more robust to occlusion and truncation. Handling symmetries in objects have posed a greater challenge to feature-based methods in part because of symmetric-induced orientation ambiguities.

\paragraph*{Keypoint-based Methods.}

% Talk about keypoint based methods and how they work for obtaining pose parameters of objects
Keypoint-based methods rely on regressing 2D keypoints of 3D objects instead of directly estimating the 3D translation and rotation. The use of keypoints as an intermediate representation of the pose and size stabilizes as well simplifies the learning problem. These methods use CNNs for feature extraction and segmentation to perform keypoint predictions through regions \cite{Hu2018_Segmentation}, heatmaps \cite{Oberweger2018_Heatmaps, Tulsiani2014_VK}, or pixel-wise predictions \cite{Capellen2019_ConvPoseCNN, Pent2018_PVNet}. Many approaches use hough voting and unit-vectors within keypoint-based methods to determine the keypoints corresponding to objects' projected 3D bounding box edges or 3D centroid point \cite{Xiang2017_PoseCNN, Pent2018_PVNet, Capellen2019_ConvPoseCNN}. While using Perspective-n-Point (PnP), these methods can obtain the pose and size from these 2D-3D correspondence keypoints.

\paragraph*{Correspondence-Mapping Methods.}
% Elaborate on correspondence mappings and how they work
Another data representation that connects 2D-3D spaces is the direct use of 2D-3D correspondence mapping of objects. Methods such as \cite{Wang2019_NOCS, Zakharov2019_DPOD, Jafari2017_iPose, Li2019_CDPN, Brachmann2014_3DObjectCoordinates} utilize dense correspondence mapping to regress intermediate representations, such as 3D object coordinates, that aid in determining the rotation and translation of the objects. \cite{Wang2019_NOCS} proposed normalized object coordinate space to integrate size estimation in 3D coordinate regression.

\paragraph*{Dense Methods.}

% Talk about dense methods and how they work
Dense methods utilize pixel-wise predictions that contribute via a reduction function to the overall object's pose. These methods have been used to regress keypoints, correspondence maps, and direct components of the pose, i.e rotation, translation, and size. Through reduction schemes, such as hough voting, RANSAC, and native averaging, dense predictions are converted into instance-wise predictions.

\subsection{Addressing the Gap in the Literature} \label{sec:literature_gap}

Our research focuses on addressing the disadvantages in the groundbreaking publication of \cite{Wang2019_NOCS} - i.e., slow inference and dependence on depth information. Their approach regressed dense correspondence mapping via a heavy MaskRCNN framework (210 ms) and later used the Umeyama algorithm (30 ms) for pose alignment. Their approach greatly benefited in performance by using correspondence mapping - yet this design decision slowed the model's speed. Our approach takes inspiration from RGB methods \cite{Pent2018_PVNet, Capellen2019_ConvPoseCNN, Xiang2017_PoseCNN} by using the smaller ResNet-FPN framework and regressing both intermediate representations and direct attributes. Through these representations, our method allows the computation of the pose and size with greater computation efficiency. By making our method only use RGB images, we make our approach compatible with most modern cameras. This delay reduction and depth independence are at a small performance penalty while rendering our method useful for time-critical and hardware-limited applications.

%%%%%%%%%%%%%%%%%%%%%%%%%%%%%%%%%%%%%%%%%%%%%%%%%%%%%%%%%%%%%%%%%%%%%%%%%%%%
\section{Methodology}
\label{ch:methodology}

\subsection{Framework Overview}

\begin{figure*}[th]
    \centering
    \input{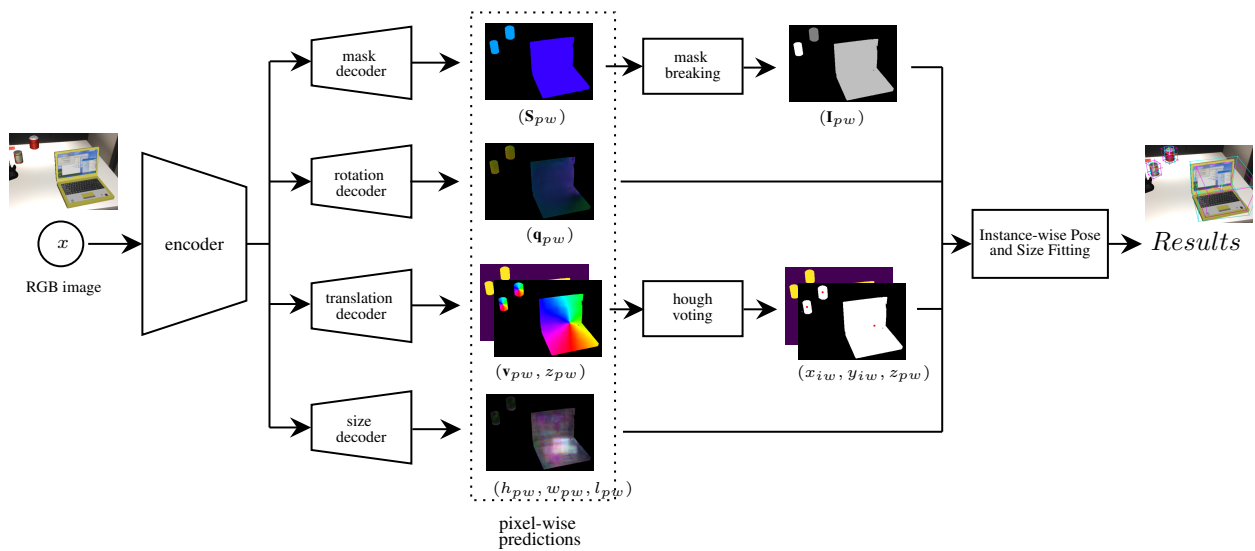}
    \caption{Model's Overall Architecture}
    \label{fig:model_overall_architecture}
\end{figure*}

% First introduce the concept of decoupled attributes and how we approach obtaining the multiple parameters involved in pose and size estimation
In our approach to 6D pose and size estimation, we decoupled the regression for each spatial component of the objects. This lead to our architecture handling the classification, rotation, translation, and size estimation in separate decoders. They are demonstrated in our architecture outline in Fig. \ref{fig:model_overall_architecture}. The decoupling of these attributes ensures stability in training and better regression as each branch effectively adapts to the typical range of the designated data.

% Give an more detail overview and description 
As shown in Fig. \ref{fig:model_overall_architecture}, our data pipeline is composed of CNN-generated pixel-wise predictions, mask breaking, hough voting, and aggregation. The input image's features are extracted by the encoder and then later used by the decoupled decoders. These decoders generate multiple dense pixel-wise predictions for segmentation, direct regression, and unit vector regression. The segmentation output is converted to instance masks via mask breaking. The unit vectors are used in hough voting to generate a centroid hypothesis for a detected object. After converting the intermediate data, aggregation takes place to match pixel-wise predictions to specific instances. In the following sections, each component of the data pipeline will be discussed in more detail.

\subsection{Pixel-Wise Predictions}

% Mention how the pixel-wise predictions are generated. Elaborate why we choose multiple decoders and what are the predictions generated in this section.
Our proposed method creates dense pixel-wise predictions for all parameters of the final pose. Our individual parameter decoders are inspired by \cite{Hu2018_Segmentation, Zakharov2019_DPOD, Xiang2017_PoseCNN, Capellen2019_ConvPoseCNN} - we noticed that small independent decoders improved stability and performance in training without a significant increase in inference time. The mask branch creates a segmentation mask. The rotation branch regresses dense pixel-wise quaternion predictions. We utilized quaternions here instead of rotation matrix due to the lower number of parameters to regress. This is to ensure that the problem space is smaller and less complex. The translation branch regressed both dense predictions for centroid unit vectors and the depth. The size branch generates dense $(h, w, l)$ predictions. 

% Elaborate the speed performance of our approach compared to other methods
By using dense pixel-wise predictions with the ResNet framework, our approach generates predictions for multiple objects in a single step - via a global context. Later in the process, the aggregation of these dense predictions is performed in an optimized batch manner - enabling the quick translation between pixel-wise to instance-wise pose and size information. Our method differs from other published works \cite{Wang2019_NOCS, Sundermeyer2019_Implicit} that perform 2D object detection and serially estimate an object's spatial attributes. Our method process multiple instances in parallel - thereby reducing the delay when multiple objects are captured in the image.

\paragraph*{Class Masking and Compression.}

% Explain how we convert class-wise data into categorical-wise data
Our method outputs pixel-wise predictions for each class to ensure the data of different classes do not interfere and lower the model's performance. Afterward, we utilize the segmentation mask to compress the pixel-wise predictions and reduce their dimensionality to match categorical data. By doing this, our model can effectively estimate the pose and size of multiple classes without any speed penalty.

\subsection{Segmentation Mask Breaking}

\begin{figure}[h]
    \centering
    \input{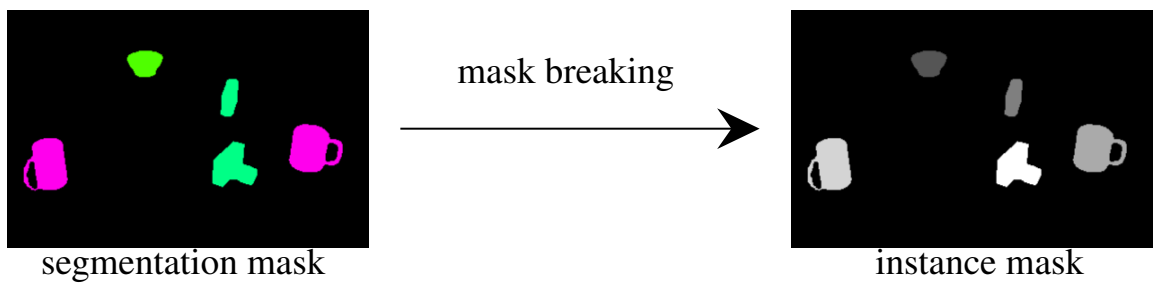}
    \caption{Mask Breaking}
    \label{fig:mask_breaking}
\end{figure}

% Explain the conversion from segmentation to instance masks in our implementation
To match the pixel-wise information between all predictions, we convert the segmentation mask to a collection of instance masks. We utilize the GPU-accelerated implementation of {scipy.ndimage.label} provided in the CuPy library \cite{Okuta2017CuPyA} to perform mask breaking. Once we convert the instance mask, it's matched with the corresponding dense pixel-wise to each object instance captured in an image by matching it with the instance mask.

% Explain why we used the segmentation mask and then the instance mask in the pipeline
Through our ResNet-FPN implementation, class segmentation and mask breaking improve the performance and shorten the training time of the model. It also allowed the optimization of the aggregation later down the pipeline. By breaking the segmentation mask into instance masks, the reduction function of the pixel-wise predictions can be performed in parallel for all instances captured in the instance masks.

\subsection{Pixel-Wise Hough Voting}

\begin{figure}[h]
    \centering
    \newcommand{\SCALE}{1.25}

\tikzset{every picture/.style={line width=0.75pt}} %set default line width to 0.75pt        

\begin{tikzpicture}[x=0.75pt,y=0.75pt,yscale=-\SCALE,xscale=\SCALE,every node/.style={scale=0.9*\SCALE}]

%Shape: Grid [id:dp2838508457063502] 
\draw  [draw opacity=0] (89,27.5) -- (290.16,27.5) -- (290.16,208.03) -- (89,208.03) -- cycle ; \draw  [color={rgb, 255:red, 65; green, 199; blue, 227 }  ,draw opacity=1 ] (89,27.5) -- (89,208.03)(109,27.5) -- (109,208.03)(129,27.5) -- (129,208.03)(149,27.5) -- (149,208.03)(169,27.5) -- (169,208.03)(189,27.5) -- (189,208.03)(209,27.5) -- (209,208.03)(229,27.5) -- (229,208.03)(249,27.5) -- (249,208.03)(269,27.5) -- (269,208.03)(289,27.5) -- (289,208.03) ; \draw  [color={rgb, 255:red, 65; green, 199; blue, 227 }  ,draw opacity=1 ] (89,27.5) -- (290.16,27.5)(89,47.5) -- (290.16,47.5)(89,67.5) -- (290.16,67.5)(89,87.5) -- (290.16,87.5)(89,107.5) -- (290.16,107.5)(89,127.5) -- (290.16,127.5)(89,147.5) -- (290.16,147.5)(89,167.5) -- (290.16,167.5)(89,187.5) -- (290.16,187.5)(89,207.5) -- (290.16,207.5) ; \draw  [color={rgb, 255:red, 65; green, 199; blue, 227 }  ,draw opacity=1 ]  ;
%Straight Lines [id:da3228798163519655] 
\draw [color={rgb, 255:red, 123; green, 123; blue, 123 }  ,draw opacity=1 ] [dash pattern={on 4.5pt off 4.5pt}]  (118.83,117.5) -- (289.16,60.69) ;
%Straight Lines [id:da32215835053859876] 
\draw [color={rgb, 255:red, 123; green, 123; blue, 123 }  ,draw opacity=1 ] [dash pattern={on 4.5pt off 4.5pt}]  (179.17,57.5) -- (179.49,207.03) ;
%Straight Lines [id:da5297798657318198] 
\draw [color={rgb, 255:red, 123; green, 123; blue, 123 }  ,draw opacity=1 ] [dash pattern={on 4.5pt off 4.5pt}]  (258.83,97.83) -- (149,207.5) ;
%Straight Lines [id:da4074622591182031] 
\draw [color={rgb, 255:red, 123; green, 123; blue, 123 }  ,draw opacity=1 ] [dash pattern={on 4.5pt off 4.5pt}]  (259.57,177.5) -- (109,27.5) ;
%Straight Lines [id:da2513870087196304] 
\draw    (118.83,117.5) -- (144.28,109.26) ;
\draw [shift={(147.14,108.34)}, rotate = 522.06] [fill={rgb, 255:red, 0; green, 0; blue, 0 }  ][line width=0.08]  [draw opacity=0] (10.72,-5.15) -- (0,0) -- (10.72,5.15) -- (7.12,0) -- cycle    ;
%Straight Lines [id:da8330774821509703] 
\draw    (179.17,57.5) -- (179.17,84.67) ;
\draw [shift={(179.17,87.67)}, rotate = 270] [fill={rgb, 255:red, 0; green, 0; blue, 0 }  ][line width=0.08]  [draw opacity=0] (10.72,-5.15) -- (0,0) -- (10.72,5.15) -- (7.12,0) -- cycle    ;
%Straight Lines [id:da7566809866401223] 
\draw    (258.83,97.83) -- (240.69,115.93) ;
\draw [shift={(238.57,118.05)}, rotate = 315.07] [fill={rgb, 255:red, 0; green, 0; blue, 0 }  ][line width=0.08]  [draw opacity=0] (10.72,-5.15) -- (0,0) -- (10.72,5.15) -- (7.12,0) -- cycle    ;
%Straight Lines [id:da19496961492777842] 
\draw    (259.57,177.5) -- (240.12,158.17) ;
\draw [shift={(238,156.05)}, rotate = 404.84000000000003] [fill={rgb, 255:red, 0; green, 0; blue, 0 }  ][line width=0.08]  [draw opacity=0] (10.72,-5.15) -- (0,0) -- (10.72,5.15) -- (7.12,0) -- cycle    ;
%Shape: Circle [id:dp32013261595305753] 
\draw  [fill={rgb, 255:red, 255; green, 0; blue, 0 }  ,fill opacity=1 ] (169.67,97.17) .. controls (169.67,91.92) and (173.92,87.67) .. (179.17,87.67) .. controls (184.41,87.67) and (188.67,91.92) .. (188.67,97.17) .. controls (188.67,102.41) and (184.41,106.67) .. (179.17,106.67) .. controls (173.92,106.67) and (169.67,102.41) .. (169.67,97.17) -- cycle ;
%Shape: Circle [id:dp4908747171107699] 
\draw  [fill={rgb, 255:red, 255; green, 0; blue, 0 }  ,fill opacity=1 ] (210,137.5) .. controls (210,132.25) and (214.25,128) .. (219.5,128) .. controls (224.75,128) and (229,132.25) .. (229,137.5) .. controls (229,142.75) and (224.75,147) .. (219.5,147) .. controls (214.25,147) and (210,142.75) .. (210,137.5) -- cycle ;
%Shape: Circle [id:dp3105458811361519] 
\draw  [fill={rgb, 255:red, 0; green, 194; blue, 255 }  ,fill opacity=1 ] (109.33,117.5) .. controls (109.33,112.25) and (113.59,108) .. (118.83,108) .. controls (124.08,108) and (128.33,112.25) .. (128.33,117.5) .. controls (128.33,122.75) and (124.08,127) .. (118.83,127) .. controls (113.59,127) and (109.33,122.75) .. (109.33,117.5) -- cycle ;
%Shape: Circle [id:dp09326729440568005] 
\draw  [fill={rgb, 255:red, 0; green, 194; blue, 255 }  ,fill opacity=1 ] (169.67,57.5) .. controls (169.67,52.25) and (173.92,48) .. (179.17,48) .. controls (184.41,48) and (188.67,52.25) .. (188.67,57.5) .. controls (188.67,62.75) and (184.41,67) .. (179.17,67) .. controls (173.92,67) and (169.67,62.75) .. (169.67,57.5) -- cycle ;
%Shape: Circle [id:dp5543978917130674] 
\draw  [fill={rgb, 255:red, 0; green, 194; blue, 255 }  ,fill opacity=1 ] (249.33,97.83) .. controls (249.33,92.59) and (253.59,88.33) .. (258.83,88.33) .. controls (264.08,88.33) and (268.33,92.59) .. (268.33,97.83) .. controls (268.33,103.08) and (264.08,107.33) .. (258.83,107.33) .. controls (253.59,107.33) and (249.33,103.08) .. (249.33,97.83) -- cycle ;
%Shape: Circle [id:dp2770929366297048] 
\draw  [fill={rgb, 255:red, 0; green, 194; blue, 255 }  ,fill opacity=1 ] (250.07,177.5) .. controls (250.07,172.25) and (254.32,168) .. (259.57,168) .. controls (264.81,168) and (269.07,172.25) .. (269.07,177.5) .. controls (269.07,182.75) and (264.81,187) .. (259.57,187) .. controls (254.32,187) and (250.07,182.75) .. (250.07,177.5) -- cycle ;

% Text Node
\draw (111.67,113.23) node [anchor=north west][inner sep=0.75pt]  [font=\tiny]  {$\mathbf{p}_{1,0}$};
% Text Node
\draw (124.67,96.69) node [anchor=north west][inner sep=0.75pt]  [font=\tiny]  {$\mathbf{v}(\mathbf{p}_{1,0})$};
% Text Node
\draw (252.67,173.85) node [anchor=north west][inner sep=0.75pt]  [font=\tiny]  {$\mathbf{p}_{2,0}$};
% Text Node
\draw (249.67,154.52) node [anchor=north west][inner sep=0.75pt]  [font=\tiny]  {$\mathbf{v}(\mathbf{p}_{2,0})$};
% Text Node
\draw (251.67,94.52) node [anchor=north west][inner sep=0.75pt]  [font=\tiny]  {$\mathbf{p}_{2,1}$};
% Text Node
\draw (171.67,53.52) node [anchor=north west][inner sep=0.75pt]  [font=\tiny]  {$\mathbf{p}_{1,1}$};
% Text Node
\draw (183.33,70.52) node [anchor=north west][inner sep=0.75pt]  [font=\tiny]  {$\mathbf{v}(\mathbf{p}_{1,1})$};
% Text Node
\draw (249,114.85) node [anchor=north west][inner sep=0.75pt]  [font=\tiny]  {$\mathbf{v}(\mathbf{p}_{2,1})$};
% Text Node
\draw (175,94.02) node [anchor=north west][inner sep=0.75pt]  [font=\tiny]  {$\mathbf{h}_{1}$};
% Text Node
\draw (215.17,134.35) node [anchor=north west][inner sep=0.75pt]  [font=\tiny]  {$\mathbf{h}_{2}$};

\end{tikzpicture}
    \caption{Hough Voting Scheme}
    \label{fig:hough_voting_scheme}
\end{figure}
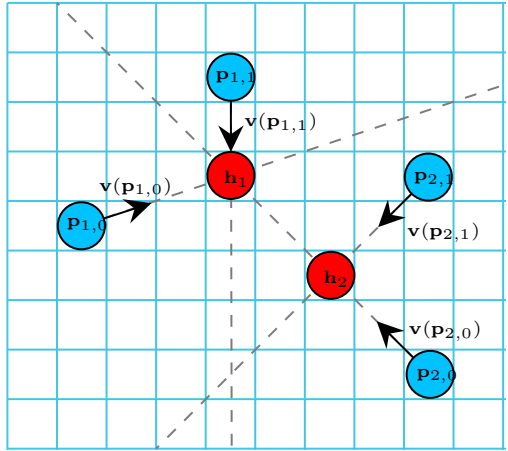

% Give credit to PVNet for borrowing parts of their implementation
For our regression of objects' $x$ and $y$ translation attributes, we used the intermediate centroid unit vectors that point towards the projection of the 3D center of an object. To convert these unit vectors to proper translation parameters, we used the popular hough voting approach. For the hough voting scheme, we adapted the CUDA-accelerated implementation proposed by PVNet \cite{Pent2018_PVNet} for our problem as it provides a fast and accurate method to process multiple centroids in a batched fashion. 

% Explain how the adapted hough voting scheme works, step by step
% Firstly, introduce how the unit vectors are constructed and defined
In our adapted hough voting algorithm, the pixel-wise centroid unit vectors are translated into a final $\textbf{c}=(u,v)^T$ hypotheses within the image plane. The centroid unit vectors $\textbf{v}(\textbf{p})$ of a pixel $\textbf{p}$ is defined by the relative location of the pixel $\textbf{p}$ from the centroid $\textbf{c}$, shown in Eq. \ref{eq:centroid_unit_vector_declaration}. With this definition, the centroid unit vectors point towards the projected 3D centroid of objects.
\begin{align} \label{eq:centroid_unit_vector_declaration}
    \textbf{v}(\textbf{p}) = \frac{\textbf{c} - \textbf{p}}{||\textbf{c} - \textbf{p}||_2}
\end{align}

% Secondly, talk about how hypotheses are constructed.
During the class masking and compression step, we apply bit-wise masking on pixel-wise unit vectors with the mask of the same classes. This step prepares the data to generate $N$ number of hypotheses. As shown in Eq. \ref{eq:hypothesis_generation_and_weighing}, we construct a hypothesis, $\textbf{h}$, for the projected centroid by obtaining the intersection between a random pair of unit vectors.
\begin{align} \label{eq:hypothesis_generation_and_weighing}
    \textbf{h}_{i} = \textbf{v}(\textbf{p}_{i,0}) \cap \textbf{v}(\textbf{p}_{i,1})
\end{align}

% Third, weight calculation for hypothesis
Once we construct $N$ number of hypotheses, we calculate the weights for each hypothesis by incorporating the rest of the object's unit vectors. These weights measure confidence in the matching hypothesis. The weights are calculated by counting the number of unit vectors that agree with the hypothesis - that is, that they point towards the hypothesis. In Eq. \ref{eq:weight_calculation}, the $\theta$ is a threshold (usually $0.99$), $O$ is the object's pertaining unit vector pixel-wise predictions. 
\begin{equation} \label{eq:weight_calculation}
    w_i = \; \sum_{\textbf{p} \in O} \boldsymbol{I} \left( \frac{(\textbf{h}_i - \textbf{p})^T}{||\textbf{h}_i - \textbf{p}||_2}\textbf{v}(\textbf{p}) \geq \theta \right)
\end{equation}

% Final pose calculation via averaging of weighted hypothesis
The final centroid hypothesis is determined by calculating the weighted average of the $N$ hypotheses. By using the hypothesis weights, the entirety of the object's unit vectors is included in the final hypothesis calculation. This contributes to faster learning with a smaller requirement of the number of hypotheses generated.
\begin{align} \label{eq:final_hypothesis}
    \textbf{h}_{final} = \frac{\sum_{i=0}^N w_i \textbf{h}_i}{\sum_{i=0}^N w_i}
\end{align}

% Talk about how we convert the final hypothesis into a translation vector
The resulting $\textbf{h}_{\text{final}}$ is later used, by combining it with the regressed depth $z$, to construct the translation vector $\textbf{t}$ of an object. To completely reconstruct the pose and size parameters of instances captured by the model, we perform an aggregation step to convert the pixel-wise predictions to instance-wise, and this part will be elaborated in the next section.

% This diagrams should truly be in experiments and results. However, if we place there it won't be placed in the right location.

\subsection{Aggregation}

\begin{figure}[h]
    \centering
    \input{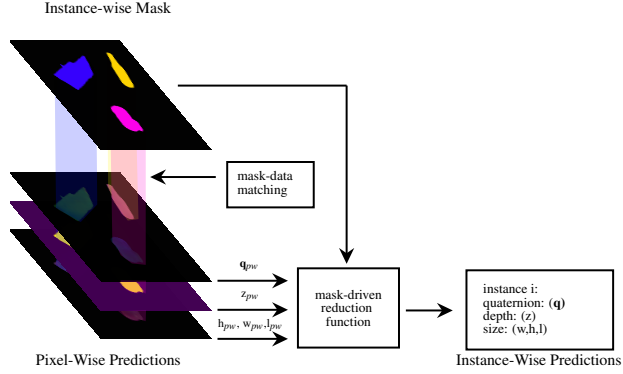}
    \caption{Aggregation via Masked Average}
    \label{fig:aggregation}
\end{figure}

% Why do we need the aggregation routine and how does it work?
With the creation of dense pixel-wise predictions for the pose and size variables, we need to compress dense predictions into instance-wise predictions to attach these parameters to captured objects. The overall aggregation routine is illustrated in Fig. \ref{fig:aggregation}. Our method utilizes the instance mask to individually extract the instance's information using a mask-drive reduction function. To ensure that our method remains real-time, we selected the fast and simple naive average of the masked dense predictions. The related reduction function is shown in Eq. \ref{eq:reduction_function} - where $\textbf{a}$ is a placeholder for any pixel-wise predictions, $O$ is the object's pixel-wise predictions, and the $I$ is the binary instance mask.
\begin{align} \label{eq:reduction_function}
    \textbf{a}_{aggregated} = \frac{\sum_{\textbf{p} \in O} \textbf{a}(\textbf{p})}{\sum_{\textbf{p} \in O} \boldsymbol{I} \left ( I(\textbf{p}) = 1)\right )} 
\end{align}

% Mention how in the aggregation step, we do not include the centroid unit vectors since these are processed by the hough voting step.
Here, we convert dense pixel-wise predictions into instance-wise predictions that allow us to create complete instance profiles that contain the translation, size, and rotation parameters. Except for the centroid unit vectors, these are not included in the aggregation routine - these pixel-wise predictions are handled by the hough voting step.

\subsection{Ground Truth and Prediction Matching}

% Discuss how the ground truth and predicted aggregated is matched to allow for training.
During training, we focus on comparing the instance-wise predictions instead of the pixel-wise predictions. It shifts the focus of the optimization problem to the estimated final pose and size parameters rather than the intermediate dense pixel-wise predictions. Therefore, we matched the ground truth instances and the predicted instances using the 2D intersection over union (IoU) metric. By calculating all of the 2D IoU's between the ground truth and predicted instance masks, we matched the instances by assigning them to their corresponding highest 2D IoU score match. We noticed how direct optimization was more stable during our training process and prevented the model's estimation performance from degrading for smaller objects - due to their smaller pixel counts.

\subsection{Loss Functions}

% Mention how we used different loss functions for each tasks
Similar to our decoupling approach, we use separate loss functions for each branch of the model. Afterward, we sum the individual contribution of each loss to determine the total. Each loss is structured to account for the range and dimension of each parameter's problem space. 
\begin{equation}
    L_{total} \coloneqqf L_{mask} + L_{sym-quat} + L_{centroid} + L_{depth} + L_{scales}
\end{equation}

\paragraph*{Segmentation.}

% Mention the loss function used for segmentation
The loss function used for segmentation is the summation between multi-class cross-entropy and focal \cite{Lin2017_FocalLoss} loss functions. 
\begin{align}
    L_{mask} \coloneqqf L_{ce}(\overline{m}, m) + L_{focal}(\overline{m}, m)
\end{align}

\paragraph*{Quaternion.}

% Explain QLoss function
For regressing the quaternion, we initially used QLoss, as shown in Eq. \ref{eq:qloss} as $L_{quat}$, referred in \cite{Billings2018_SilhoNet}. This loss function accounts for the internal symmetry of quaternions. However, it does not account for the symmetric object's axis of symmetry. We propose $L_{sym-quat}$ for making the quaternion loss function fit for symmetric objects.
\begin{align} \label{eq:qloss}
    L_{quat} \coloneqqf \log( \epsilon +1-|\mathbf{\overline{q}} \cdot \mathbf{q} |)
\end{align}

\paragraph*{Symmetric Object Handling.}

\begin{figure}[h]
    \centering
    \includegraphics[width=0.3\columnwidth]{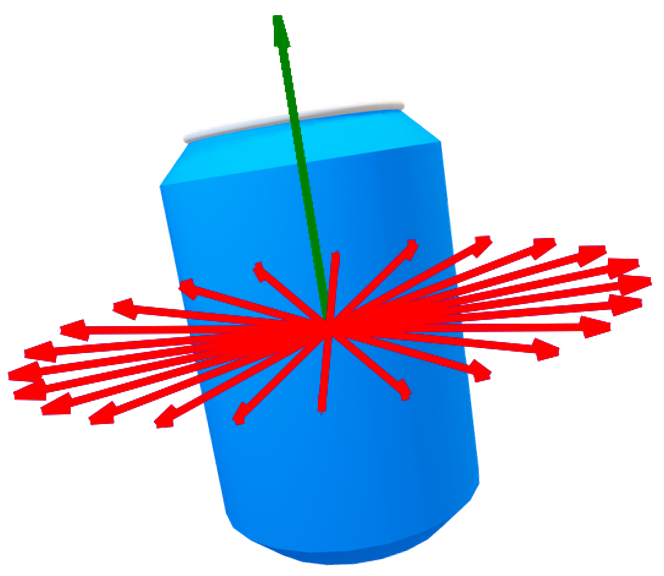}
    \caption{Quaternion Axis-of-Symmetry Rotation}
    \label{fig:quaternion_symmetry}
\end{figure}

\begin{figure*}[th]
    \centering
    \subfloat[FastPoseCNN Performance Curves]{\includegraphics[width=0.49\textwidth]{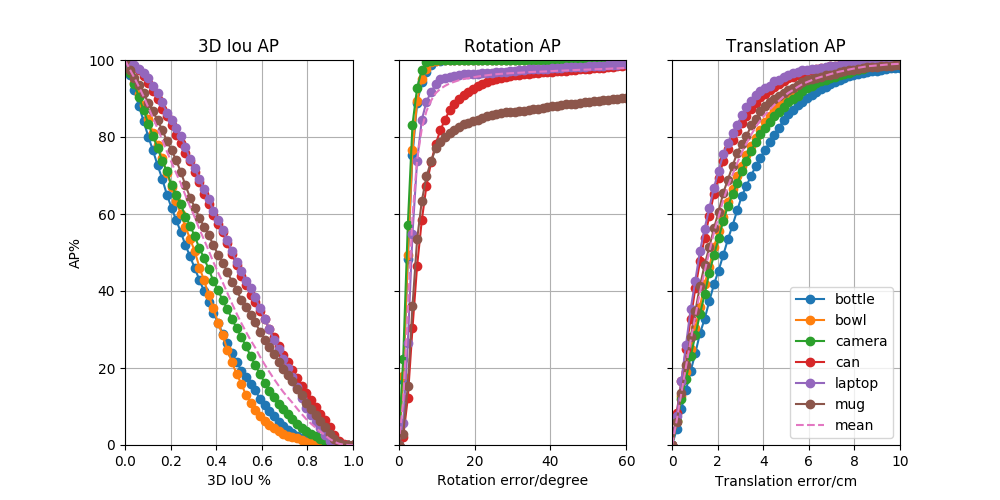}}
    \subfloat[NOCSNet Performance Curves]{\includegraphics[width=0.49\textwidth]{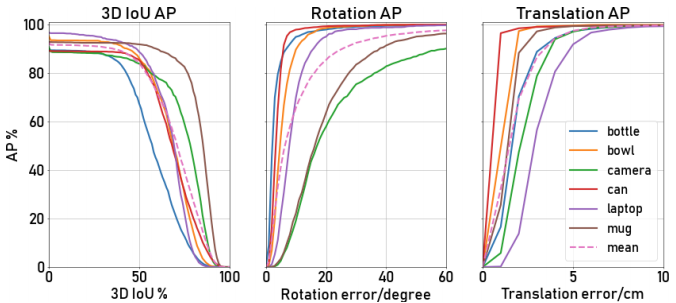}}
    \caption{3D detection and 6D pose estimation results for CAMERA validation}
    \label{fig:CAMERA_mAP}
\end{figure*}

\begin{figure*}[th]
    \centering
    \subfloat[FastPoseCNN Performance Curves]{\includegraphics[width=0.49\textwidth]{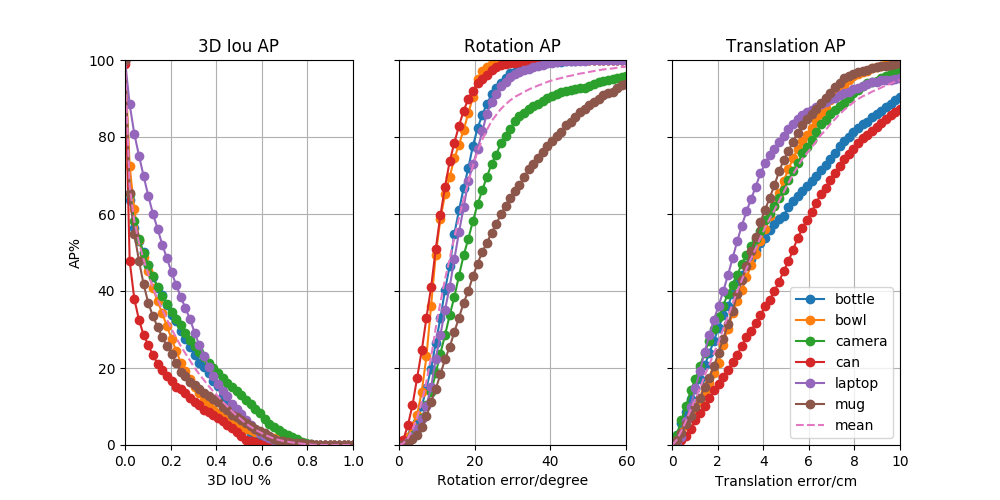}}
    \subfloat[NOCSNet Performance Curves]{\includegraphics[width=0.49\textwidth]{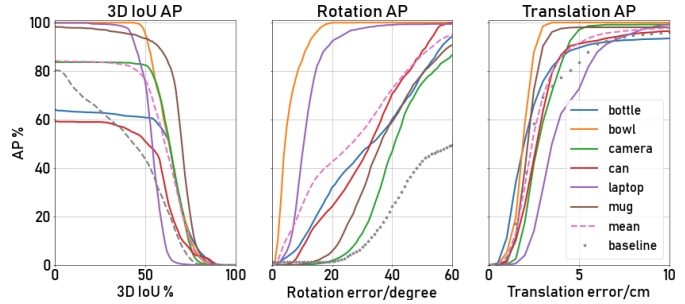}}
    \caption{3D detection and 6D pose estimation results for REAL test dataset}
    \label{fig:REAL_mAP}
\end{figure*}

% Explain how we defined the symmetric quaternion loss function
A major issue when performing rotation estimation is symmetry. Many of the objects present in the CAMERA/REAL datasets have an axis of symmetry. We explored and determined that if we directly regressed the rotation without considering the axis of symmetric of certain objects, the model yielded drastically lower performance for those objects. 

% Pictorially explain how we handle the symmetry of objects
As shown in Fig. \ref{fig:quaternion_symmetry}, our strategy is to generate a set of equivalent ground truth quaternions rotated around the axis of symmetric. The goal is to capture all possible correct quaternion orientations and concentrate the quaternion optimization problem for symmetric objects to capture the axis of symmetry.

% Explain how we construct valid ground truth quaternion through our processes
Therefore, we annotate the axis of symmetry for each type of symmetric object. We specify a set of rotation angles, $\theta = \{0^{\circ}, 1^{\circ}, ..., 359^{\circ}\}$, which we use to form a new set of transformation quaternion, $\hat{\textbf{q}}_i = \{\hat{\textbf{q}}_{0^{\circ}}, ..., \hat{\textbf{q}}_{|\theta|}\}$. By applying the transformation quaternion to the ground truth quaternion $\overline{\textbf{q}}_i = \hat{\textbf{q}}_i \overline{\textbf{q}} \hat{\textbf{q}}^{-1}_i$, we construct ground truth quaternions, $\overline{\textbf{q}}_i = \{\overline{\textbf{q}}_{0^{\circ}}, ..., \overline{\textbf{q}}_{|\theta|}\}$ that capture all possible valid rotations. As shown in Eq. \ref{eq:sym-quat}, we calculate the $L_{quat}$ for each $\overline{\textbf{q}}_i$ and use the lowest loss value. 
\begin{align} \label{eq:sym-quat}
    L_{sym-quat} \coloneqqf 
    \begin{cases}
    \min_{i=i,...,|\theta|} L_{quat}\left(\mathbf{\overline{\mathbf{q}}}_{i} ,\mathbf{q}\right) & \text{if symmetric} \\
    L_{quat}\left(\overline{\textbf{q}}, \textbf{q}) \right) & \text{otherwise}
    \end{cases}
\end{align}

\paragraph*{Centroid Unit Vectors.}

% Explain the loss function used for the centroid and why
Early in the training phase, large quantities of outliers are generated by the untrained hough voting scheme. Through multiple trials, we determined that L1 outperformed smooth L1 and L2 in this task. Therefore, for regressing the projected 3D centroid, we use L1 loss on each coordinate space of the centroid.
\begin{align}
    L_{centroid} \coloneqqf \ell _{1}(\mathbf{\overline{c}} |_{x} ,\mathbf{c} |_{x}) +\ell _{1}(\mathbf{\overline{c}} |_{y} ,\mathbf{c} |_{y})
\end{align}

\paragraph*{Depth.}

% Explain the complexity of the depth and how we log(z) to improve stability.
The ambiguity of estimating the depth makes this component of the model have the lowest performance. Additionally, minor errors in the depth estimation have significant negative effects on metrics such as 3D IoU thresholds. We follow the stabilizing technique proposed by \cite{Capellen2019_ConvPoseCNN} of estimating the $\log(z)$ instead of the $z$. This enhances the performance and stabilizes the training of the model.
\begin{align}
    L_{depth} \coloneqqf \ell _{1}(\log(\overline{z}) ,\log( z))
\end{align}

\paragraph*{Size Scales.}

% Explain how the sizes of the object's were regressed
For regressing the size of objects, we use L1 loss functions on each component of the size. This ensures the outliers in scales parameters were better handled compared to using smooth L1 or L2 loss functions.
\begin{align}
    L_{scales} \coloneqqf \ell _{1}(\overline{\mathbf{s}} |_{h} ,\mathbf{s} |_{h}) +\ell _{1}(\overline{\mathbf{s}} |_{w} ,\mathbf{s} |_{w}) +\ell _{1}(\overline{\mathbf{s}} |_{l} ,\mathbf{s} |_{l})
\end{align}

%%%%%%%%%%%%%%%%%%%%%%%%%%%%%%%%%%%%%%%%%%%%%%%%%%%%%%%%%%%%%%%%%%%%%%%%%%%%
\section{Experiments and Results}
\label{ch:experimentsResults}

\begin{table*}[th]
    \centering
    \subfloat[Validation on CAMERA25 \label{tab:camera_performance}]{
\begin{tabular}{c c c}
     \hline
     & \multicolumn{2}{c}{CAMERA} \\
     \hline
     methods & NOCS & OURS \\
     \hline
     $3\text{D}_{25}$ & 90.13 & 66.69 \\
     $3\text{D}_{50}$ & 87.58 & 32.33 \\
     $5^{\circ} \& 5\text{cm}$ & 38.14 & 67.16 \\
     $10^{\circ} \& 5\text{cm}$ & 61.24 & 84.14 \\
     $10^{\circ} \& 10\text{cm}$ & 62.01 & 91.02 \\
     
\end{tabular}}
    \hfill
    \subfloat[Class-Wise performance on CAMERA25 \label{tab:classwise_camera_performance}]{
\begin{tabular}{c | c c c c c c}
     \hline
     Class & \multicolumn{3}{c}{OURS} & \multicolumn{3}{c}{NOCS} \\
     & $3\text{D}_{50}$ & $5^{\circ}$\&$5\text{cm}$ & $10^{\circ}$\&$10\text{cm}$ & $3\text{D}_{50}$ & $5^{\circ}$ \& $5\text{cm}$ & $10^{\circ}$\&$10\text{cm}$ \\
     \hline
     bottle & 20.22 & 77.15 & 97.88 & 89.18 & 78.99 & 91.56\\
     bowl   & 35.08 & 79.18 & 99.13 & 91.36 & 51.18 & 87.77\\
     can    & 26.97 & 81.19 & 99.24 & 85.28 & 78.78 & 97.32\\
     laptop & 54.12 & 71.07 & 93.45 & 85.56 & 16.52 & 63.67\\
     camera & 39.53 & 51.88 & 76.90 & 83.59 & 2.01  & 17.47\\ 
     mug    & 37.00 & 44.03 & 79.53 & 90.53 & 13.50 & 14.28\\
\end{tabular}}
    \caption{Additional breakdown comparison information in CAMERA.}
    \label{fig:tables}
\end{table*}

\subsection{Tools and Source Code}

% Explain tools and source code that we borrowed to make our research possible
Our experiments and model were implemented using the PyTorch framework \cite{Paszke2019_PyTorch} and PyTorch-Lightning \cite{falcon2019_pytorch-lightning} open source libraries. We also used PyTorch Segmentation \cite{Yakubovskiy2019_SegmantationModelsPyTorch} pre-trained FPN-resnet18 model to jump start our training. Our model adapted PVNet's CUDA-accelerated hough voting scheme \cite{Pent2018_PVNet}.

\subsection{Implementation and Training}

% List implementation information
We initialize the ResNet18-FPN backbone with the weights pre-trained on Imagenet. In the first stage of training, we focus on training the mask branch of the model. We used a batch size of 2, an initial learning rate of $1 \times 10^{-4}$, and a RAdam optimizer with a weight decay of $3 \times 10^{-4}$. We disabled the hough voting, mask breaking, and aggregation steps that are later used in the second stage of training. After freezing all the layers except those found in the encoder and mask branch, we begin the speed-optimized training for 50 epochs. In the second stage of training, we trained the entire model for another 50 epochs. We used the same hyperparameters for this stage while enabling all intermediate steps of the model during training. The learning rate is reduced by a factor of $0.25$ by a plateau scheduler in both stages. 

\subsection{Metrics}

% Provide a description of the metrics and possibly a definition
We adopt the metrics used in \cite{Wang2019_NOCS} to evaluate our results. These metrics include the 3D IoU and the mean average precision (mAP) where the error is less than \textit{m} cm for translation and $\textit{n}^{\circ}$ for rotation. By considering separate metrics for translation, rotation, and scale, we can more clearly present the performance of the model. For symmetric objects, we apply the same technique used in the symmetric loss function by rotating the ground truth quaternion and 3D bounding box by the axis-of-symmetric and selecting the highest performance value. 

\subsection{Comparison to SOTA Methods}

% Begin our comparison to SOTA Methods
As the following, we report and compare our category-level results to the reported values of NOCSNet \cite{Wang2019_NOCS} on both datasets. Due to our training structure, comparisons with NOCSNet for the CAMERA25 validation dataset use the reported data with the same training amount and type.

% Example Output of the model
% \input{Figures/experiments_and_results/images/example_output}

% Talk about NOCS CAMERA DATASET
\paragraph*{CAMERA25 Validation.}
We tested our model against the CAMERA25 validation set after only training with the 275K CAMERA training set. Our model achieves \textbf{32.33\%} for 3D IoU at 50\% and an $5^{\circ} \& 5$cm mAP of \textbf{66.69\%}. The precision curves for these metrics are shown in Fig. \ref{fig:CAMERA_mAP}. Note that these metrics are quite challenging because of the perspective ambiguities introduced by the unknown depth of objects.

% Talk about REAL test dataset
\paragraph*{REAL275 Testing.}
After training the model on the CAMERA275 training dataset, we train an additional 50 epochs specifically on the REAL training dataset. When tested on the REAL test dataset, our model performs \textbf{7.18\%} for 3D IoU at 50\% and an $5^{\circ} \& 5$cm mAP of \textbf{5.18\%}. The precision curves for these metrics are shown in Fig. \ref{fig:REAL_mAP}. The shifting of domains is especially tough when real data is not sufficient. Both NOCSNet and our proposed method have significantly decreased performance when testing on the REAL test dataset.

% Give an reflection on these results.
The approach proposed by \cite{Wang2019_NOCS} yields higher performance for the 3D IoU metric primarily because of its high-quality correspondence mapping method used to regress the object's scales and its use of depth images. However, correspondence mapping methods, like NOCSNet, do not handle well axis of symmetry. Our results verify this - as our direct regression rotation method achieves higher mAP. It is also important to consider that our method regresses the depth with good accuracy yet it negatively affects the 3D IoU metric. Our method achieves good performance while not requiring additional depth information. 

% Provide more numerical information about our methods
In Table \ref{fig:tables}, we present our per-class performance in \ref{tab:classwise_camera_performance} and a further breakdown in performance in \ref{tab:camera_performance}. Through the class-wise information, it is clear that our method can capture the symmetric nature of objects better compared to \cite{Wang2019_NOCS} using our $L_{quat-sym}$ loss function.

\subsection{Inference \& Time-Breakdown}

% Talk about our speed performance
The primary purpose of this research is to provide a real-time monocular version of \cite{Wang2019_NOCS}. Our method can achieve an average delay of 43ms (23 fps) during inference - allowing our method to be considered real-time for pose estimation applications. A run-time breakdown is presented in Table \ref{tab:model_time_breakdown}. As far as we know, we are the first to propose a monocular category-level pose and size estimation framework with a real-time inference.

\begin{table}[h]
    \centering
    
\begin{tabular}{l l}
\hline 
 Component & Delay Time (ms) \\
\hline 
 Feature Extractor & 18.570 \\
 Aggregation & 4.808 \\
 Hough Voting & 12.894 \\
 RT Calculation & 3.769 \\
 Class Compression & 2.660 \\
 Total & 43.355 (23fps) \\
\end{tabular}
    \caption{Total Model Time-Breakdown}
    \label{tab:model_time_breakdown}
\end{table}

\subsection{Limitations}

% Discuss limitations to our research
In this section, we elaborate on the limitations of FastPoseCNN. First, the Hough Voting algorithm obtained from PVNet requires a CUDA-compatible GPU device to run it. Second, to use a different camera, additional training would be required. As referred by the pose estimation community, the camera intrinsics were baked into the model's parameters when we trained on the CAMERA and REAL datasets. This should not affect performance drastically, but it should be noticed. Third, the pose and size estimated by FastPoseCNN are excellent to a certain degree. FastPoseCNN should not be used for applications that depend on precision-critical objects' pose and size estimation. The research presented here is a proof-of-concept and would require further research and development to become a commercially reliable system.

%%%%%%%%%%%%%%%%%%%%%%%%%%%%%%%%%%%%%%%%%%%%%%%%%%%%%%%%%%%%%%%%%%%%%%%%%%%%
\section{Conclusion}
\label{ch:conclusion}

In this thesis, we have introduced a real-time monocular category-level pose and size estimation framework that globally detects and estimates an object's pose and size parameters via dense pixel-wise predictions. FastPoseCNN is excellent at estimating the pose of symmetric objects while running in real-time. We showed how our specialized $L_{sym-quat}$ loss function improves the training of the model and outperforms NOCSNet in rotation estimation. Our experiment and analysis section demonstrates the performance and speed of FastPoseCNN compared to previous works. In future work, we plan on creating a more robust intermediate size and depth interpretation to achieve higher performance in 3D IoU and translation offset metrics. Another possible future research direction includes the use of the NVIDIA TensorRT library \cite{nvidia2021_TensorRT} to further accelerate the model and increase its throughput. 

%%%%%%%%%%%%%%%%%%%%%%%%%%%%%%%%%%%%%%%%%%%%%%%%%%%%%%%%%%%%%%%%%%%%%%%%%%%%
\section*{Acknowledgments}

%Bibliography
\bibliographystyle{unsrt}  
\bibliography{references}  

\end{document}